# Geared Rotationally Identical and Invariant Convolutional Neural Network Systems

*ShihChung B. Lo, Ph.D.[1,2,3], Matthew T. Freedman, M.D.[2,3], Seong K. Mun, Ph.D.[2], and Heang-Ping Chan, Ph.D.[4]

[1] Radiology Department, Georgetown University Medical Center, Washington DC 20007
[2] Arlington Innovation Center: Health Research, Virginia Tech, Arlington, VA 22203
[3] Oncology Department, Georgetown University Medical Center, Washington DC 20007
[4] Radiology Department, University of Michigan Medical Center, Ann Arbor, Michigan, 48109
*e-mail: <dcben0@gmail.com> or <benlo@vt.edu>

## Abstract

Theorems and techniques to form different types of transformationally invariant processing and to produce the same output quantitatively based on either transformationally invariant operators or symmetric operations have recently been introduced by the authors. In this study, we further propose to compose a geared rotationally identical CNN system (GRI-CNN) with a small step angle by connecting networks of participated processes at the first flatten layer. Using an ordinary CNN structure as a base, requirements for constructing a GRI-CNN include the use of either symmetric input vector or kernels with an angle increment that can form a complete cycle as a "gearwheel". Four basic GRI-CNN structures were studied. Each of them can produce quantitatively identical output results when a rotation angle of the input vector is evenly divisible by the step angle of the gear. Our study showed when an input vector rotated with an angle does not match to a step angle, the GRI-CNN can also produce a highly consistent result. With a design of using an ultra-fine gear-tooth step angle (e.g., $1^o$ or $0.1^o$), all four GRI-CNN systems can be constructed virtually isotropically.

## 1. Introduction

Construction of rotationally invariant convolutional neural networks system has been attempted for applications in medical image pattern recognition since its early development (Lo 1993; 1995(a); 1995(b)). Besides the use of rotated and reflected input vectors as a part of data augmentation, circular path convolutional neural networks system was also designed to analyze object possessing circular and radial correlation such as breast lesions on mammogram (Lo 1998; 2002). Recently several CNN structures constructed with transformation components were proposed by many investigators. Among them, methods to achieve rotationally invariant have been proposed by Gens and Domingos (2014); Dieleman et al. (2015; 2016); Cohen et al. (2016); Cheng et al (2016); Ravanbakhsh et al. (2017); Zaheer et al. (2017b); Guttenberg et al. (2016); Cohen et al. (2017); Cohen et al. (2018), and Weiler et al. (2018). Fourier Transform for a rotationally invariant CNN was studied by Marcos et al. (2016) and Chidester et al (2018). Follmann et al. proposed to incorporate rotational invariance into the feature-extraction part (2018). Li et al proposed to use cycle, isotonic, and decycle layers to speed up processes with rotated filters (2018). However, none of these approaches was able to produce quantitatively identical result when an input is transformed. Recently, both theorems and techniques to form translationally or rotationally invariant processing and to produce the same output quantitatively based on either transformationally invariant operators or symmetric operations have been introduced by this research team (Lo 2018(a) and 2018(b)). Specifically, three transformationally identical convolution neural network systems can be constructed by either (1) operating with a set of self transformationally invariant kernels, (2) processing through a set of symmetric operations, and (3) combining the input vector with its rotated and/or reflected versions. In this study, we further propose to construct isotropic CNN systems that can produce either no or negligible difference in the output layer with unlimited input vector versions of rotation.

## 2. Methods

In this study, we focused on properties of matrix rotation within the CNN. Hence, only the rotationally identical property and systems possessing the rotationally identical property would be our study subjects without considering other forms of transformation.

The optimal goal of this study was intended to construct rotationally identical (RI) systems when the input vector is rotated with an arbitrary angle. This can be approached by composing newly developed rotationally identical CNN systems using rotated versions of the input vector without involving an interpolation (type-1 transformation). In the same time, rotated versions of the input vector involving interpolation (type-2 transformation) would be used to form other groups of RI processes. By regulating the type-2 transformations and combining all participated processes at the first flatten layer or at the input layer, a composed RI-CNN can be constructed. Specifically, our strategies for constructing a comprehensive RI-CNN system with a small step angle are:

a) using type-1 transformations of the input vector that inherently include in an RI-CNN process;

b) using type-2 transformations of the input vector and their RI counterparts to produce other groups of RI-CNN processes.

c) connecting all sets of participated type-2 transformations at the input or the first flatten layer to form a comprehensive RI-CNN system covering small and large angles of rotations.

### 2.1. Review of Rotationally identical CNN (RI-CNN) systems

#### 2.1.1. An RI-CNN (RI-CNN-1) constructed by a set of self rotationally invariant kernels

A typical CNN processes input vector with series of convolutional processes, non-linear function and network processing to produce a resultant vector on the output layer. With a set of rotationally invariant kernels used in convolution layers, shared weights corresponding to their transformation element positions at the first flatten layer can be arranged with the same symmetric pattern. These weights would perform inner product with intermediate resultant matrix on the last convolution/pooling layer. Each node on the first flatten layer would then be fully connected to nodes on the following layer and finally reach to the output layer. Having a CNN operated by this manner, the output of the CNN would be identical with the same family of rotationally identical kernels for all corresponding transformations of input vector (Vi). Hence it is a rotationally identical CNN (RI-CNN) when it possesses a set of RI kernels on all convolution layers and weights sharing on the first flatten layer. RI kernel refers to a kernel satisfying $K_h = Rs\{ K_h \}$, where Rs{.} is the pre-defined rotationally symmetric transform. “h” denotes each of convolution layers or the first flatten layer. Such an RI-CNN can be considered as a function:

$$\text{RI-CNN-1}[ Vi ] = \text{RI-CNN-1}[ R_s\{ Vi \} ] = Vo_1 \qquad ...(1)$$

Therefore, there is no need to use any corresponding rotated versions of the input vector to enter into the input layer of an RI-CNN as they all end-up the same result at each node on and after the first flatten layer.

#### 2.1.2. An RI-CNN (RI-CNN-2) constructed by combining input vector of the same rotationally symmetric versions

A rotationally identical CNN (RI-CNN-2) using non-symmetric convolution kernels can also be constructed by using a set of symmetric operations or combined input vector of the rotationally symmetric versions. Mathematically speaking, this type of rotationally identical CNN is related but different from the first type of rotationally identical CNN (RI-CNN-1). The RI-CNN-2 has two equivalent forms: either combining grouped data before or at the end of convolutions.

RI-$CNN_{\sum_{r=1}^{M} RI\{K\}}$[ Vi ] = RI-$CNN_{\sum_{r=1}^{M} RI\{K\}}$[ $RI$\{Vi\} ]

= RI-$CNN_{RI\{K\}}[\sum_{r=1}^{M} RI\{Vi\}]$ = $Vo_2$ . …(2)

Note that the outputs received from RI-CNN-1 and RI-CNN-2 are different.

2.2. Composition of regulated rotation versions of the input vector or kernels to form a complete cycle geared RI-CNN (GRI-CNN) system

A conventional data augmentation method is used to increase the training sample and hopes to achieve a better generalization performance when a similar data pattern is presented. However, the RI-CNN system is designed to perform an unequivocal test result with a rotated input vector. Since the dihedral symmetry of order 8 (Dih4) covers all other rotation symmetry groups, let's use it for the development of GRI-CNN structure as an example. With the RI-CNN as a base, each type-2 transformation with $x^o$ rotation as a step angle satisfying (x mod 90) ≠ 0 and (90 mod x) = 0, can be used to perform another set of RI-CNN sub-channel. A total of "m" (m·x = 90) RI-CNN sub-channels must be used to form a complete cycle geared RI-CNN (GRI-CNN) system with each RI-CNN sub-channel as a "tooth" of the whole cycle gear. The GRI-CNN as a function can be expressed as

GRI-CNNm:$x^o$^[Vi] = GRI-CNNm:$x^o$^[$R_{n \cdot x^o}$\{Vi\}] ...(3)

where m:$x^o$^ denotes rotation step at an angle increment of $x^o$ with a total "m" number of steps to from a GRI-CNN. $R_{n \cdot x^o}$\{.\} is the rotation function with an angle of $n \cdot x^o$, and n = 1,2, ..., m-1.

There are multiple ways to construct a GRI-CNN that can produce RI results with a small step angle design. Four basic GRI-CNN systems were designed and are reported below:

(1) Sharing all symmetric kernels among all rotation step sub-channels

By sharing RI kernels at each convolution layer, "m" versions of rotated input vector are combined in the input layer with a step angle of $x^o$ to cover $90^o$ rotation followed by an RI-CNN-1 process. Fig. 1 shows a schematic diagram of a GRI-CNN/SSK system using a set of shared symmetric kernels.

(2) Sharing all non-symmetric kernels among all rotation step sub-channels

By sharing non-symmetric kernels at each convolution layer, all (8·m in 2D and 24·m in 3D) versions of rotated input vector are combined in the input layer with a step angle of $x^o$ to meet the requirement of Dih4 symmetry followed by an ordinary CNN process. Fig. 2 shows a schematic diagram of a GRI-CNN/SNK system using a set of shared non-symmetric kernels.

(3) Sharing symmetric kernels within each group of angles possessing the Dih4 symmetry

By sharing symmetric kernels within each Dih4 group (step angles meet Dih4 symmetry criterion) but not sharing with other groups, followed by "m" sub-channels of convolution processes (each represents a gear-tooth group) for the construction of the 3rd type of GRI-CNN system. Fig. 3 shows a schematic diagram of a GRI-CNN/GSK system using a set of shared symmetric kernels within each Dih4 group.

(4) Sharing non-symmetric kernels within each group of angles possessing the Dih4 symmetry

By sharing non-symmetric kernels within each Dih4 group but not sharing with other groups, multiple (8 in 2D and 24 in 3D) versions of rotated input vector are combined in the input layer followed by "m" pipelines of convolution processes (each represents a gear-tooth group) for the construction of 4th type of GRI-CNN system. Fig. 4 shows a schematic diagram of a GRI-CNN/GNK system using a set of shared non-symmetric kernels within each Dih4 group.

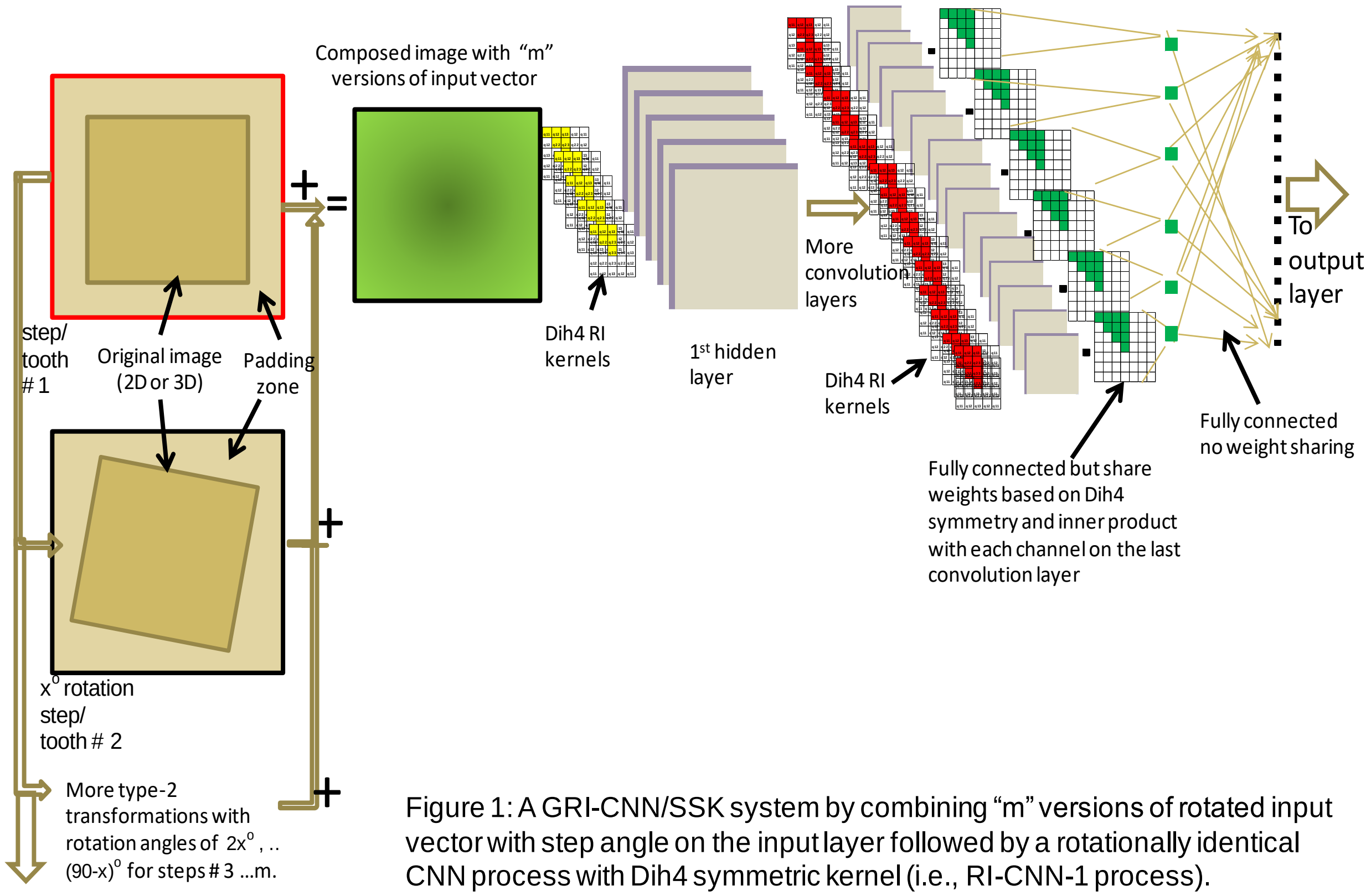

Figure 1: A GRI-CNN/SSK system by combining "m" versions of rotated input vector with step angle on the input layer followed by a rotationally identical CNN process with Dih4 symmetric kernel (i.e., RI-CNN-1 process).

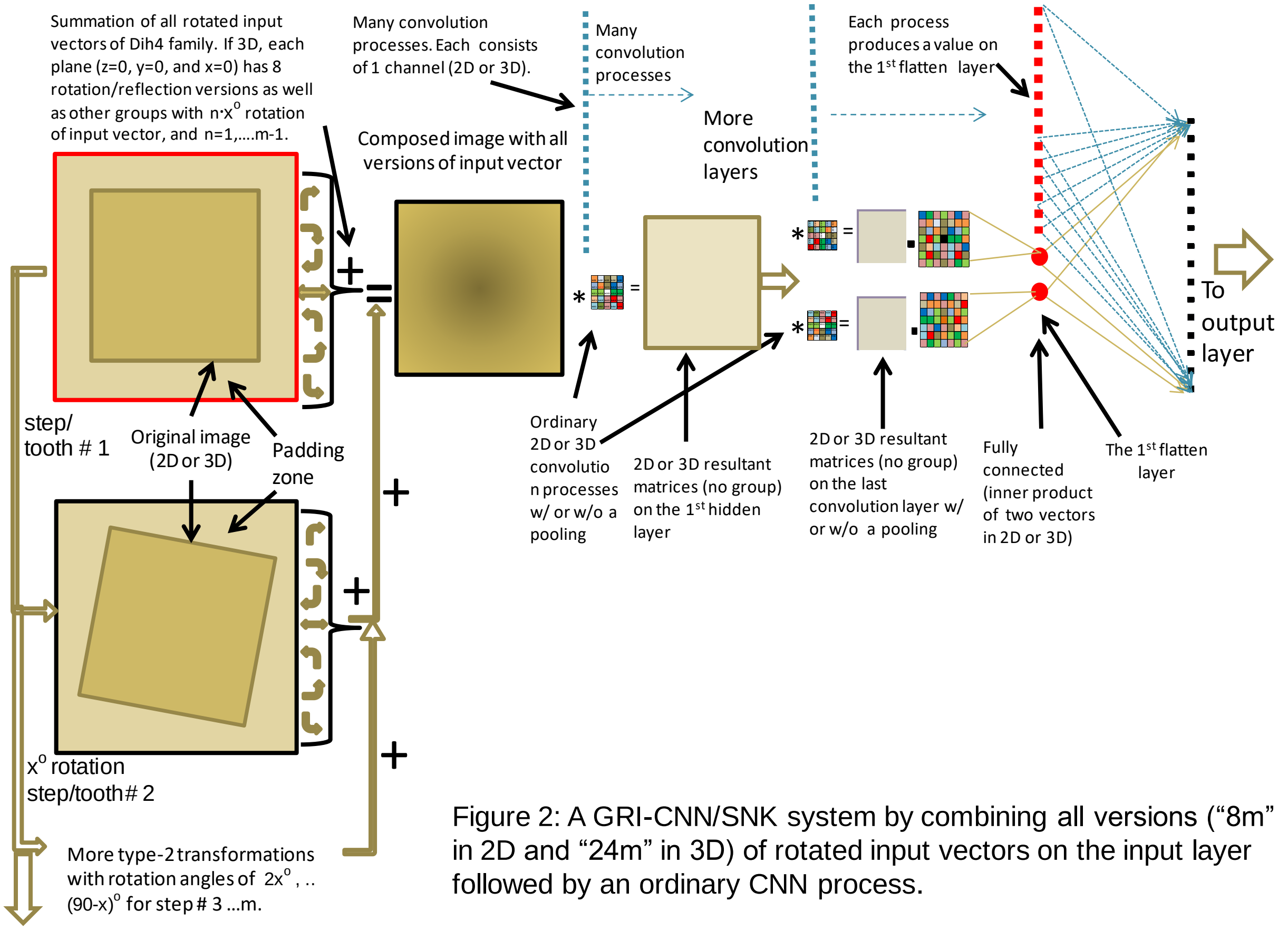

Figure 2: A GRI-CNN/SNK system by combining all versions ("8m" in 2D and "24m" in 3D) of rotated input vectors on the input layer followed by an ordinary CNN process.

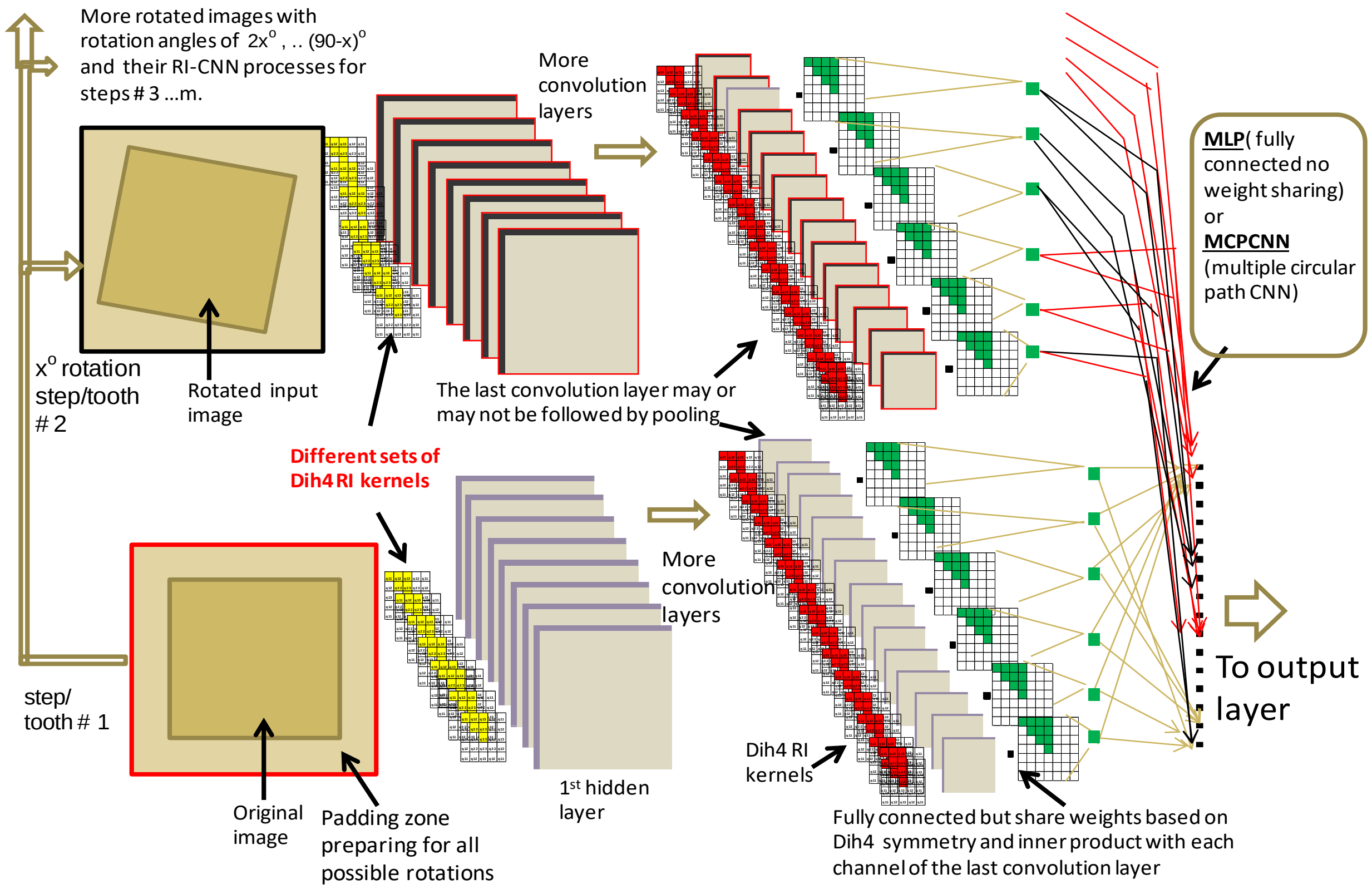

Figure 3: A GRI-CNN/GSK system consists of multiple RI-CNNs associated with corresponding rotated versions of the same input vector.

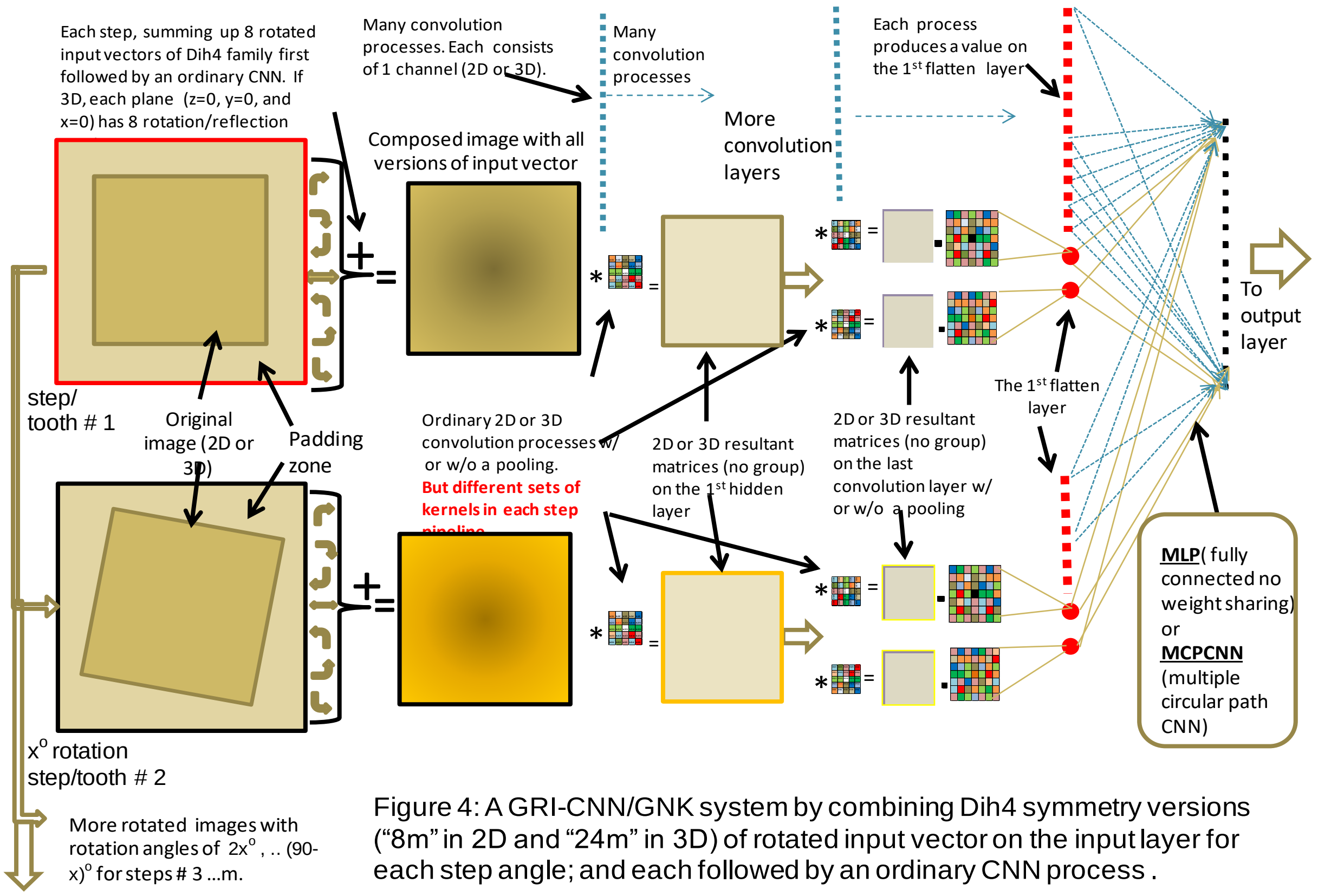

Figure 4: A GRI-CNN/GNK system by combining Dih4 symmetry versions ("8m" in 2D and "24m" in 3D) of rotated input vector on the input layer for each step angle; and each followed by an ordinary CNN process .

Assuming a GRI-CNN/GSK or a GRI-CNN/GNK system and its RI-CNN sub-channels all end at the first flatten layer, the result of GRI-CNN is the combined result of all RI-CNN sub-channels:

$$\text{GRI-CNN}_{m:x^o}\text{\^{}}[\ V_i\ ] = \sum_{n=0}^{m-1} (\ \text{RI-CNN}[\ R_{n\cdot x^o}\{\ V_i\ \}\ ] \quad ...(4)$$

These 4 types of GRI-CNN systems follow an important theorem that any convolutional process with or without pooling is commutative with combining transformations either before or at the end of process if carrying the same type of rotationally symmetric kernels. Before the second flatter layer, each RI-CNN performs independently. For the GRI-CNN/SSK and the GRI-CNN/GSK systems, the combining process takes place at the input layer before entering the convolution process section of the CNN. Since all kernels are shared across all gear-tooth groups, GRI-CNN/SSK or GRI-CNN/GSK structure would automatically establish its own rotationally identical system.

For the implementation of composing a complete cycle GRI-CNN, a padding zone can be added as an extension for the original image, so that any rotated version of original image requiring an interpolation would be geometrically covered in the same size of a square region in 2D and a cubic region in 3D, respectively as shown on the left side of Figs. 1-4.

Each set of RI kernels of the same rotation family (i.e., kernel elements are Dih4 symmetric) should be consistently applied to all convolutions along each sub-channel until the last convolution layer and would still possess the same RI property. One important operation to from an RI-CNN system is that the node values computed on the first flatten layer from all RI-CNN sub-systems basically are connected as a fan-in or specially arranged to connect to nodes on the next flatten layer and eventually merge into each node on the output layer. This approach would generate a GRI-CNN system inclusive of participated gear-tooth groups with type-2 rotations of the input vector. There are two conditions to be met for the design of $x^o$ rotation as a step angle (i.e., $m \cdot x = 90$, and “m” must be an integer), any participated rotation angle can be used as the starting angle paralleled with its partners (i.e., the starting angle + $n \cdot x^o$, for n=1,2, ..., m-1). With this design, any composed GRI-CNN system with the circular sequence arrangement would possess the same groups and each has the same RI-CNN sub-channels like a "full cycle gearwheel". Similar approaches can be employed to construct all 4 types of GRI-CNN systems in 3D as long as an increment solid angle can form a "full cycle gear-sphere".

With a fully connected networks after the first flatten layer, the composed networks can be processed as a typical multiple layer perceptron (MLP). However, since feature banks at the first flatten layer are computed from RI-CNN sub-channels in the last two GRI-CNN structures (i.e., GSK and GNK) which are circularly and spherically correlated in 2D and 3D, respectively, the use of a multiple circular path convolutional neural networks (MCPCNN) would be more suitable for training this part of networks [Lo 1998; 2002]. With this method, corresponding computed features are arranged as a circular data set and would be processed by a 1D symmetric kernel with a circular path manner. Neither a beginning nor an ending point can be defined. The MCPCNN technique was initiated by Lo et al for lung and breast lesion research. In addition, several training methods can be taken to train the GRI-CNN composed from "m" RI-CNN sub-channels.

(a) Train one RI-CNN at a time and merge their feature banks in the second run of the training.

(b) Train several RI-CNN sub-channels together and merge all feature banks in the second run of the training.

(c) Train all RI-CNN sub-channels together.

(d) Instead of training from scratch for each or partially combined sub-channel, transfer training can be used from one training to another and to all combined sub-channels.

It will take some time for investigators in various fields to find out the performance of the proposed GRI-CNN systems and their variants.  In this study, we performed several preliminary studies to investigate the RI property of GRI-CNN structures and their output consistency in terms of achieving no or negligible difference when a fine step (gear-tooth) angle was designed.

### 3. Experiments and Results

We used 1,000 breast lesion images cropped from collected mammograms. Of them, 500 lesions were malignant, and 500 lesions were benign. Radiologically speaking, the diagnostic determination would be the same for each lesion despite of any rotated and reflected image version. The size of cropped mammographic lesion images was formatted with a matrix size of 128 x 128 pixels. Fig. 5 and 6 show samples of mammographic lesions.

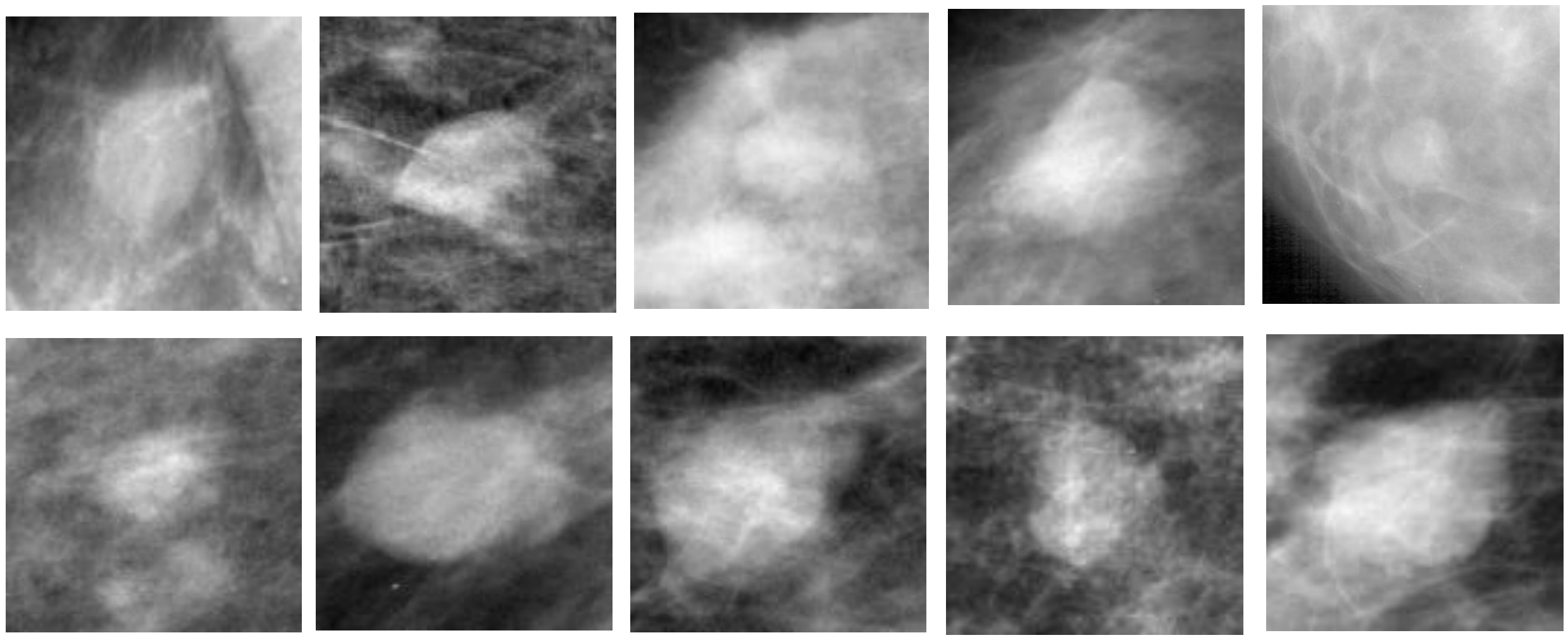

Figure 5. Samples of benign mammographic lesions

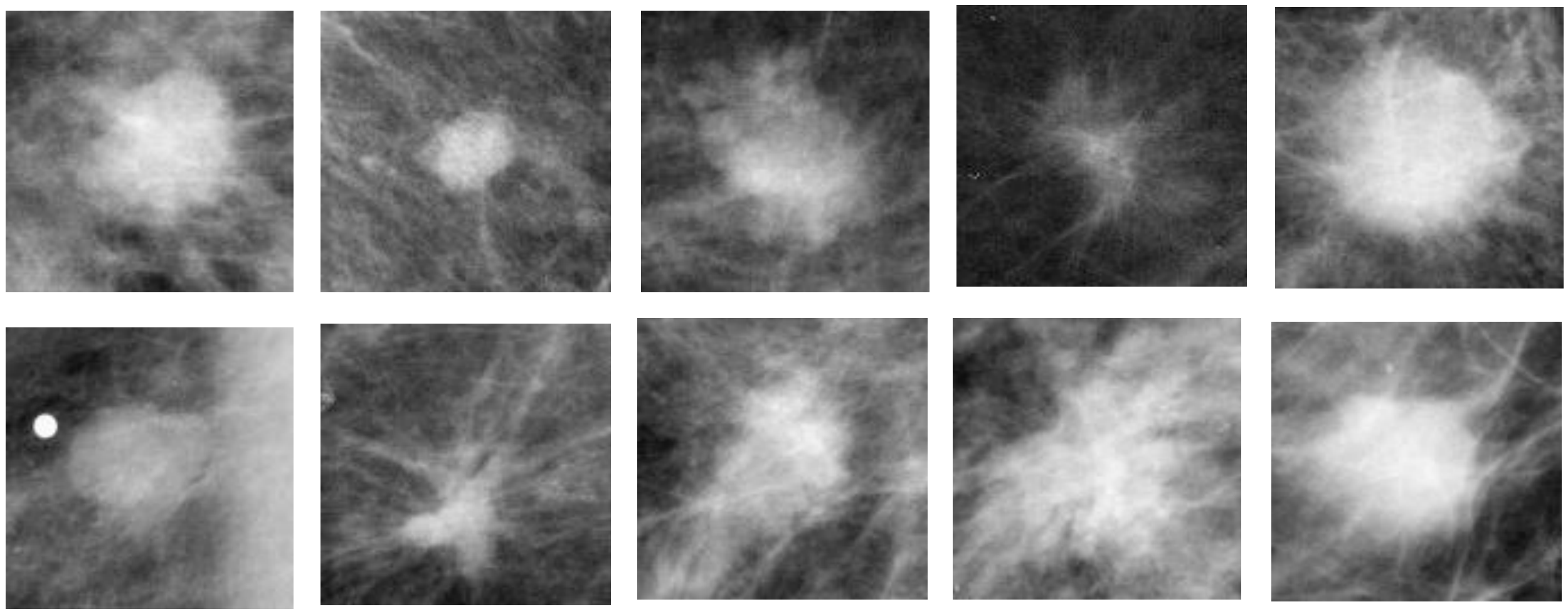

Figure 6. Samples of malignant mammographic lesions

Our experiment was performed as follows.

1) Each GRI-CNN consisted of 5 convolution hidden layers and one fatten layer. Each signal came out from a convolution process was modulated by the sigmoid function and was compared with the baseline (i.e., combined results of the first gear-tooth and its Dih4 symmetry counterparts).

2) Each rotated input vector and non-shared convolution kernels used in GSK and GNK were rotated accordingly through the bilinear interpretation from their corresponding original ones.

3) Convolution kernel size was randomly assigned between 3x3 and 11x11. The size of kernel connecting between the resultant matrix size on the last convolution layer and each node on the flatten layer depends on the size of the last resultant matrix.

4) The angle between adjacent "gear teeth" to form a GRI-CNN system was set at $1^o$ (i.e., the step angle $x^o=1^o$). The incremental angle for each test was 1/10 (i.e., $0.1^o$) of the step angle.

5) All output data to be examined were normalized by the output data obtained using the original input vector and its Dih4 symmetry counterparts as the baseline.

6) Each GRI-CNN structure was tested using 100,000 sets of parameters.

7) For each test, an angle was specified with $(i \cdot 90^o + n \cdot x^o + f \cdot x^o/10))$ where "f" = 0,1,2...10; "n" is a random integer between 0 and m-1 (m = 90/x); and "i" is a random integer between 0 and 3. In addition, each rotated input vector for the test was either reflected or not reflected randomly.

8) Each weight coefficient was initially filled with a random number between -0.5 and 0.5 to simulate the initialization of a CNN training. The study (100,000 tests) using all kernels filled with random numbers representing worst scenarios.

9) Three hundred CNN training settings with various kernel size arrangement were used to train for the classification of 1,000 mammographic lesions. For kernels filled with random numbers, the results are shown in Tables I, 2, 3, and 4 for GRI-CNN systems with SSK, SNK, GSK, and GNK structures, respectively.

10) After several epochs (~ 50), the kernels extracted from epochs 50 to 500 were used for the second part of the study. They were evaluated with same settings stated above to examine the output consistency of each GRI-CNN system. The results are shown in Tables 5 - 8.

Special symbols used in the tables are defined as: RR - random choice of reflection or no reflection. RAN{.} - random integer number generator bounded by a pre-defined range.

Table 1. Output consistency of the GRI-CNN/**SSK** whose kernels filled with random numbers

| Test angle | mean | maximum difference | minimum difference | MSE | Error <1% | <2 % | <3 % | <4 % | <5 % | >5 % |
|---|---|---|---|---|---|---|---|---|---|---|
| $RR\{ RAN\{I\} \cdot 90^o + RAN\{n\} \cdot 1^o \}$ | 1 | N/A | N/A | N/A | | | | | | |
| $RR\{ RAN\{I\} \cdot 90^o + RAN\{n\} \cdot 1^o + 0.1^o \}$ | 1 | 0.01517 | -0.01786 | 0 | 99997 | 3 | 0 | 0 | 0 | 0 |
| $RR\{ RAN\{I\} \cdot 90^o + RAN\{n\} \cdot 1^o + 0.2^o \}$ | 1 | 0.03592 | -0.01646 | 0 | 99996 | 2 | 0 | 2 | 0 | 0 |
| $RR\{ RAN\{I\} \cdot 90^o + RAN\{n\} \cdot 1^o + 0.3^o \}$ | 1 | 0.04696 | -0.01207 | 0 | 99996 | 2 | 1 | 0 | 1 | 0 |
| $RR\{ RAN\{I\} \cdot 90^o + RAN\{n\} \cdot 1^o + 0.4^o \}$ | 1 | 0.03586 | -0.01385 | 0.0000001 | 99992 | 4 | 2 | 2 | 0 | 0 |
| $RR\{ RAN\{I\} \cdot 90^o + RAN\{n\} \cdot 1^o + 0.5^o \}$ | 1 | 0.03823 | -0.01646 | 0.0000001 | 99990 | 6 | 2 | 2 | 0 | 0 |
| $RR\{ RAN\{I\} \cdot 90^o + RAN\{n\} \cdot 1^o + 0.6^o \}$ | 1 | 0.05554 | -0.01610 | 0.0000001 | 99989 | 9 | 1 | 0 | 0 | 1 |
| $RR\{ RAN\{I\} \cdot 90^o + RAN\{n\} \cdot 1^o + 0.7^o \}$ | 1 | 0.08100 | -0.02178 | 0.0000001 | 99988 | 6 | 4 | 1 | 0 | 1 |
| $RR\{ RAN\{I\} \cdot 90^o + RAN\{n\} \cdot 1^o + 0.8^o \}$ | 1 | 0.03819 | -0.02399 | 0.0000001 | 99986 | 8 | 4 | 2 | 0 | 0 |
| $RR\{ RAN\{I\} \cdot 90^o + RAN\{n\} \cdot 1^o + 0.9^o \}$ | 1 | 0.04730 | -0.02400 | 0.0000001 | 99990 | 4 | 2 | 1 | 3 | 0 |
| $RR\{ RAN\{I\} \cdot 90^o + RAN\{n\} \cdot 1^o + 1^o \}$ | 1 | 0.00004 | -0.00002 | 0 | 100000 | 0 | 0 | 0 | 0 | 0 |

Table 2. Output consistency of the GRI-CNN/**SNK** whose kernels filled with random numbers

| Test angles | mean | maximum difference | minimum difference | MSE | Error <1% | <2 % | <3 % | <4 % | <5 % | >5 % |
|---|---|---|---|---|---|---|---|---|---|---|
| $RR\{ RAN\{l\} \cdot 90^o + RAN\{n\} \cdot 1^o \}$ | 1 | N/A | N/A | N/A | | | | | | |
| $RR\{ RAN\{l\} \cdot 90^o + RAN\{n\} \cdot 1^o + 0.1^o\}$ | 1 | 0.06255 | -0.07341 | 0.0000005 | 99929 | 44 | 8 | 12 | 3 | 4 |
| $RR\{ RAN\{l\} \cdot 90^o + RAN\{n\} \cdot 1^o + 0.2^o\}$ | 1 | 0.07057 | -0.09607 | 0.0000011 | 99883 | 64 | 26 | 8 | 9 | 9 |
| $RR\{ RAN\{l\} \cdot 90^o + RAN\{n\} \cdot 1^o + 0.3^o\}$ | 1 | 0.09642 | -0.09124 | 0.0000013 | 99862 | 70 | 34 | 15 | 7 | 12 |
| $RR\{ RAN\{l\} \cdot 90^o + RAN\{n\} \cdot 1^o + 0.4^o\}$ | 1 | 0.09242 | -0.12374 | 0.0000018 | 99839 | 79 | 40 | 13 | 11 | 18 |
| $RR\{ RAN\{l\} \cdot 90^o + RAN\{n\} \cdot 1^o + 0.5^o\}$ | 1 | 0.11306 | -0.13120 | 0.0000020 | 99843 | 73 | 31 | 25 | 12 | 16 |
| $RR\{ RAN\{l\} \cdot 90^o + RAN\{n\} \cdot 1^o + 0.6^o\}$ | 1 | 0.11012 | -0.09509 | 0.0000018 | 99836 | 81 | 44 | 12 | 10 | 17 |
| $RR\{ RAN\{l\} \cdot 90^o + RAN\{n\} \cdot 1^o + 0.7^o\}$ | 1 | 0.12175 | -0.12051 | 0.0000018 | 99849 | 77 | 27 | 22 | 14 | 11 |
| $RR\{ RAN\{l\} \cdot 90^o + RAN\{n\} \cdot 1^o + 0.8^o\}$ | 1 | 0.11329 | -0.12650 | 0.0000014 | 99880 | 64 | 26 | 15 | 3 | 12 |
| $RR\{ RAN\{l\} \cdot 90^o + RAN\{n\} \cdot 1^o + 0.9^o\}$ | 1 | 0.06026 | -0.08841 | 0.0000005 | 99924 | 49 | 17 | 6 | 1 | 3 |
| $RR\{ RAN\{l\} \cdot 90^o + RAN\{n\} \cdot 1^o + 1^o\}$ | 1 | 0.00005 | -0.00003 | 0 | 100000 | 0 | 0 | 0 | 0 | 0 |

Table 3. Output consistency of the GRI-CNN/**GSK** whose kernels filled with random numbers

| Test angle | mean | maximum difference | minimum difference | MSE | Error <1% | <2 % | <3 % | <4 % |
|---|---|---|---|---|---|---|---|---|
| $RR\{ RAN\{l\} \cdot 90^o + RAN\{n\} \cdot 1^o \}$ | 1 | N/A | N/A | N/A | | | | |
| $RR\{ RAN\{l\} \cdot 90^o + RAN\{n\} \cdot 1^o + 0.1^o \}$ | 1 | 0.01517 | -0.00785 | 0 | 99998 | 2 | 0 | 0 |
| $RR\{ RAN\{l\} \cdot 90^o + RAN\{n\} \cdot 1^o + 0.2^o \}$ | 1 | 0.02234 | -0.01323 | 0 | 99998 | 1 | 1 | 0 |
| $RR\{ RAN\{l\} \cdot 90^o + RAN\{n\} \cdot 1^o + 0.3^o \}$ | 1 | 0.02645 | -0.01089 | 0 | 99997 | 2 | 1 | 0 |
| $RR\{ RAN\{l\} \cdot 90^o + RAN\{n\} \cdot 1^o + 0.4^o \}$ | 1 | 0.03277 | -0.01320 | 0 | 99995 | 2 | 2 | 1 |
| $RR\{ RAN\{l\} \cdot 90^o + RAN\{n\} \cdot 1^o + 0.5^o \}$ | 1 | 0.03034 | -0.01867 | 0.0000001 | 99993 | 4 | 2 | 1 |
| $RR\{ RAN\{l\} \cdot 90^o + RAN\{n\} \cdot 1^o + 0.6^o \}$ | 1 | 0.02897 | -0.01578 | 0 | 99996 | 3 | 1 | 0 |
| $RR\{ RAN\{l\} \cdot 90^o + RAN\{n\} \cdot 1^o + 0.7^o \}$ | 1 | 0.02689 | -0.02178 | 0 | 99996 | 2 | 2 | 0 |
| $RR\{ RAN\{l\} \cdot 90^o + RAN\{n\} \cdot 1^o + 0.8^o \}$ | 1 | 0.02319 | -0.01478 | 0 | 99998 | 1 | 1 | 0 |
| $RR\{ RAN\{l\} \cdot 90^o + RAN\{n\} \cdot 1^o + 0.9^o \}$ | 1 | 0.01345 | -0.01425 | 0 | 99999 | 2 | 0 | 0 |
| $RR\{ RAN\{l\} \cdot 90^o + RAN\{n\} \cdot 1^o + 1^o \}$ | 1 | 0.00002 | -0.00001 | 0 | 100000 | 0 | 0 | 0 |

Table 4. Output consistency of the GRI-CNN/**GNK** whose kernels filled with random numbers

| Test angles | mean | maximum difference | minimum difference | MSE | Error <0.1% | <2 % | <3 % | <4 % | <5 % |
|---|---|---|---|---|---|---|---|---|---|
| $RR\{ RAN\{l\} \cdot 90^o + RAN\{n\} \cdot 1^o \}$ | 1 | N/A | N/A | N/A | | | | | |
| $RR\{ RAN\{l\} \cdot 90^o + RAN\{n\} \cdot 1^o + 0.1^o\}$ | 1 | 0.01391 | -0.02228 | 0.0000001 | 99988 | 7 | 5 | 0 | 0 |
| $RR\{ RAN\{l\} \cdot 90^o + RAN\{n\} \cdot 1^o + 0.2^o\}$ | 1 | 0.02595 | -0.03268 | 0.0000002 | 99971 | 18 | 5 | 6 | 0 |
| $RR\{ RAN\{l\} \cdot 90^o + RAN\{n\} \cdot 1^o + 0.3^o\}$ | 1 | 0.03367 | -0.02212 | 0.0000003 | 99940 | 42 | 12 | 6 | 0 |
| $RR\{ RAN\{l\} \cdot 90^o + RAN\{n\} \cdot 1^o + 0.4^o\}$ | 1 | 0.04167 | -0.02941 | 0.0000005 | 99911 | 66 | 12 | 10 | 1 |
| $RR\{ RAN\{l\} \cdot 90^o + RAN\{n\} \cdot 1^o + 0.5^o\}$ | 1 | 0.04304 | -0.03236 | 0.0000006 | 99916 | 49 | 17 | 16 | 2 |
| $RR\{ RAN\{l\} \cdot 90^o + RAN\{n\} \cdot 1^o + 0.6^o\}$ | 1 | 0.02726 | -0.03457 | 0.0000004 | 99910 | 66 | 20 | 4 | 0 |
| $RR\{ RAN\{l\} \cdot 90^o + RAN\{n\} \cdot 1^o + 0.7^o\}$ | 1 | 0.01810 | -0.03408 | 0.0000003 | 99946 | 38 | 11 | 5 | 0 |
| $RR\{ RAN\{l\} \cdot 90^o + RAN\{n\} \cdot 1^o + 0.8^o\}$ | 1 | 0.01622 | -0.02392 | 0.0000001 | 99976 | 17 | 7 | 0 | 0 |
| $RR\{ RAN\{l\} \cdot 90^o + RAN\{n\} \cdot 1^o + 0.9^o\}$ | 1 | 0.01238 | -0.01035 | 0.0000001 | 99998 | 2 | 0 | 0 | 0 |
| $RR\{ RAN\{l\} \cdot 90^o + RAN\{n\} \cdot 1^o + 1^o\}$ | 1 | 0.00001 | -0.00001 | 0 | 100000 | 0 | 0 | 0 | 0 |

Table 5. Output consistency of the GRI-CNN/**SSK** with partially trained convolution kernel sets

| Test angles | mean | maximum difference | minimum difference | MSE | Error <0.5% | Error <1% |
|---|---|---|---|---|---|---|
| RR{ RAN{l} · $90^o$ + RAN{n} · $1^o$ } | 1 | N/A | N/A | N/A | N/A | N/A |
| RR{ RAN{l} · $90^o$ + RAN{n} · $1^o$ + $0.1^o$} | 1 | 0.00128 | -0.00495 | 0 | 100000 | 0 |
| RR{ RAN{l} · $90^o$ + RAN{n} · $1^o$ + $0.2^o$} | 1 | 0.00220 | -0.00509; -0.00144 | 0 | 99999 | 1 |
| RR{ RAN{l} · $90^o$ + RAN{n} · $1^o$ + $0.3^o$} | 1 | 0.00315 | -0.00604; -0.00443 | 0 | 99999 | 1 |
| RR{ RAN{l} · $90^o$ + RAN{n} · $1^o$ + $0.4^o$} | 1 | 0.00292 | -0.00883; -0.00211 | 0 | 99999 | 1 |
| RR{ RAN{l} · $90^o$ + RAN{n} · $1^o$ + $0.5^o$} | 1 | 0.00419 | -0.00877; -0.00263 | 0 | 99999 | 1 |
| RR{ RAN{l} · $90^o$ + RAN{n} · $1^o$ + $0.6^o$} | 1 | 0.00449 | -0.00364 | 0 | 100000 | 0 |
| RR{ RAN{l} · $90^o$ + RAN{n} · $1^o$ + $0.7^o$} | 1 | 0.00578; 0,00192 | -0.00511; -0.00244 | 0 | 99998 | 2 |
| RR{ RAN{l} · $90^o$ + RAN{n} · $1^o$ + $0.8^o$} | 1 | 0.00542; 0.00213 | -0.00387 | 0 | 99999 | 1 |
| RR{ RAN{l} · $90^o$ + RAN{n} · $1^o$ + $0.9^o$} | 1 | 0.00644; 0.00224 | -0.00669; -0.00112 | 0 | 99998 | 2 |
| RR{ RAN{l} · $90^o$ + RAN{n} · $1^o$ + $1^o$} | 1 | 0.00003 | -0.00002 | 0 | 100000 | 0 |

Table 6. Output consistency of the GRI-CNN/**SNK** with partially trained convolution kernel sets

| Test angles | mean | maximum differnce | minimum differnce | MSE | Error <0.5% | Error <= 0.1012% |
|---|---|---|---|---|---|---|
| RR{ RAN{l} · $90^o$ + RAN{n} · $1^o$ } | 1 | N/A | N/A | N/A | | |
| RR{ RAN{l} · $90^o$ + RAN{n} · $1^o$ + $0.1^o$} | 1 | 0.00076 | -0.00228 | 0 | 100000 | 0 |
| RR{ RAN{l} · $90^o$ + RAN{n} · $1^o$ + $0.2^o$} | 1 | 0.00187 | -0.00487 | 0 | 100000 | 0 |
| RR{ RAN{l} · $90^o$ + RAN{n} · $1^o$ + $0.3^o$} | 1 | 0.00214 | -0.00659; -0.00320 | 0 | 99999 | 1 |
| RR{ RAN{l} · $90^o$ + RAN{n} · $1^o$ + $0.4^o$} | 1 | 0.00209 | -0.01012; -0.00621 | 0 | 99998 | 2 |
| RR{ RAN{l} · $90^o$ + RAN{n} · $1^o$ + $0.5^o$} | 1 | 0.00208 | -0.00779; -0.00413 | 0 | 99999 | 1 |
| RR{ RAN{l} · $90^o$ + RAN{n} · $1^o$ + $0.6^o$} | 1 | 0.00222 | -0.00875; -0.00499 | 0 | 99999 | 1 |
| RR{ RAN{l} · $90^o$ + RAN{n} · $1^o$ + $0.7^o$} | 1 | 0.00302 | -0.00875; -0.00842 | 0 | 99998 | 2 |
| RR{ RAN{l} · $90^o$ + RAN{n} · $1^o$ + $0.8^o$} | 1 | 0.00273 | -0.00497 | 0 | 100000 | 0 |
| RR{ RAN{l} · $90^o$ + RAN{n} · $1^o$ + $0.9^o$} | 1 | 0.00150 | -0.00161 | 0 | 100000 | 0 |
| RR{ RAN{l} · $90^o$ + RAN{n} · $1^o$ + $1^o$} | 1 | 0.00003 | -0.00002 | 0 | 100000 | 0 |

Table 7. Output consistency of the GRI-CNN/**GSK** with partially trained convolution kernel sets

| Test angles | mean | maximum difference | minimum difference | MSE | Error <0.5% |
|---|---|---|---|---|---|
| RR{ RAN{l} · $90^o$ + RAN{n} · $1^o$ } | 1 | N/A | N/A | N/A | |
| RR{ RAN{l} · $90^o$ + RAN{n} · $1^o$ + $0.1^o$} | 1 | 0.00109 | -0.00191 | 0 | 100000 |
| RR{ RAN{l} · $90^o$ + RAN{n} · $1^o$ + $0.2^o$} | 1 | 0.00091 | -0.00225 | 0 | 100000 |
| RR{ RAN{l} · $90^o$ + RAN{n} · $1^o$ + $0.3^o$} | 1 | 0.00125 | -0.00348 | 0 | 100000 |
| RR{ RAN{l} · $90^o$ + RAN{n} · $1^o$ + $0.4^o$} | 1 | 0.00339 | -0.00436 | 0 | 100000 |
| RR{ RAN{l} · $90^o$ + RAN{n} · $1^o$ + $0.5^o$} | 1 | 0.00437 | -0.00478 | 0 | 100000 |
| RR{ RAN{l} · $90^o$ + RAN{n} · $1^o$ + $0.6^o$} | 1 | 0.00394 | -0.00389 | 0 | 100000 |
| RR{ RAN{l} · $90^o$ + RAN{n} · $1^o$ + $0.7^o$} | 1 | 0.00317 | -0.00268 | 0 | 100000 |
| RR{ RAN{l} · $90^o$ + RAN{n} · $1^o$ + $0.8^o$} | 1 | 0.00243 | -0.00138 | 0 | 100000 |
| RR{ RAN{l} · $90^o$ + RAN{n} · $1^o$ + $0.9^o$} | 1 | 0.00168 | -0.00121 | 0 | 100000 |
| RR{ RAN{l} · $90^o$ + RAN{n} · $1^o$ + $1^o$} | 1 | 0.00003 | -0.00001 | 0 | 100000 |

Table 8. Output consistency of the GRI-CNN/**GNK** with partially trained convolution kernel sets

| Test angles | mean | maximum difference | minimum difference | MSE | Error <0.5% | Error < 1% |
|---|---|---|---|---|---|---|
| $RR\{ RAN\{I\} \cdot 90^{o} + RAN\{n\} \cdot 1^{o} \}$ | 1 | N/A | N/A | N/A | | |
| $RR\{ RAN\{I\} \cdot 90^{o} + RAN\{n\} \cdot 1^{o} + 0.1^{o}$ | 1 | 0.00128 | -0.00339 | 0 | 100000 | 0 |
| $RR\{ RAN\{I\} \cdot 90^{o} + RAN\{n\} \cdot 1^{o} + 0.2^{o}$ | 1 | 0.00246 | -0.00485 | 0 | 100000 | 0 |
| $RR\{ RAN\{I\} \cdot 90^{o} + RAN\{n\} \cdot 1^{o} + 0.3^{o}$ | 1 | 0.00341 | -0.00439 | 0 | 100000 | 0 |
| $RR\{ RAN\{I\} \cdot 90^{o} + RAN\{n\} \cdot 1^{o} + 0.4^{o}$ | 1 | 0.00394 | -0.00678 | 0 | 99999 | 1 |
| $RR\{ RAN\{I\} \cdot 90^{o} + RAN\{n\} \cdot 1^{o} + 0.5^{o}$ | 1 | 0.00138 | -0.00377 | 0 | 100000 | 0 |
| $RR\{ RAN\{I\} \cdot 90^{o} + RAN\{n\} \cdot 1^{o} + 0.6^{o}$ | 1 | 0.00130 | -0.00581 | 0 | 99999 | 1 |
| $RR\{ RAN\{I\} \cdot 90^{o} + RAN\{n\} \cdot 1^{o} + 0.7^{o}$ | 1 | 0.00137 | -0.00530 | 0 | 99999 | 1 |
| $RR\{ RAN\{I\} \cdot 90^{o} + RAN\{n\} \cdot 1^{o} + 0.8^{o}$ | 1 | 0.00103 | -0.00181 | 0 | 100000 | 0 |
| $RR\{ RAN\{I\} \cdot 90^{o} + RAN\{n\} \cdot 1^{o} + 0.9^{o}$ | 1 | 0.00048 | -0.00048 | 0 | 100000 | 0 |
| $RR\{ RAN\{I\} \cdot 90^{o} + RAN\{n\} \cdot 1^{o} + 1^{o} \}$ | 1 | 0.00000 | 0.00000 | 0 | 100000 | 0 |

These results showed when the step (gear-tooth) angle set at $1^{o}$, either no difference or a few greater than 1% of errors were observed from the entire CNN training section including the use of kernels filled with random numbers. During the training, the kernels were getting more regulated and no error was found greater than 1% except a few events using the GRI-CNN/GNK structure. Normalization with the baseline output value was taken for the purposes of comparing output values obtained from its off-set counterparts. A drawback of the normalization is when the baseline value is very small, a tiny difference would be greatly amplified. Almost all normalized errors greater than 0.5% were found with a baseline output value less than 0.1. In other words, noticeable normalized errors, though very small number of events, occasionally observed when a GRI-CNN outputs a very small value. This study result was performed with kernel sets extracted from the CNN training process.

It seemed that GRI-CNN with GSK and GNK base structures performed the best output results. However, comparison must consider two facts: (1) A very high percentage of the sub-channels produced no difference or very small differences. Their composed result would behave as a group or their averaging and would result in less variability. (2) Since multiple convolution channels are commonly used in a CNN, the use of multiple SSK and GSK channels to reach similar number of free parameters in a GRI-CNN system may produce similar output consistency to that of using the GSK or the GNK as the base structure.

Additional tests were also performed using finer step angles, more hidden convolution layers and different activation functions. We found that a less difference was obtained by using finer step angles (e.g., $0.5^{o}$) or more hidden convolution layers. However, the difference was greater by using a rectified linear unit (ReLU) for the activation function as compared to that of sigmoid function. The effect was less pronounced when a leaky rectified linear unit (LReLU) was used.

In fact, the 11th test angle (the last row on each of above 8 tables) represents composed result of all step angles randomly selected from all "gear-teeth". Only very tiny small differences were observed occasionally in nearly 1 million tests with all range of parameters. These very small differences might be due to multiple steps of CNN computation and the use of bilinear interpolation for matrix rotations in our computer program implementation. Hence our study results firmly confirmed that quantitatively identical output results were obtained in all steps (gear-teeth) regardless the use of GRI-CNN structure, number of convolution layer, activation function, and with or without stride and/or pooling (should the operation be symmetrically performed).

## 4. Conclusions and Discussion

In this study, we demonstrated that a total of 4 GRI-CNN structures can be constructed using two types of rotationally identical CNN structures: RI-CNN-1 and RI-CNN-2. The former is based on symmetric operators of Dih4 symmetry and the latter is based on combining rotated input vectors with $90^o$ rotation increment and their reflections.

Again, two out of these 4 GRI-CNN structures are constructed based on RI-CNN-1:
SSK: Sharing symmetric kernels for all sub-channels.
GSK: Sharing symmetric kernels for angles within the Dih4 symmetry family as a group.

The other two are based on RI-CNN-2:
SNK: Sharing non-symmetric kernels for all sub-channels.
GNK: Sharing non-symmetric kernels for angles within the Dih4 symmetry family as a group.

The RI-CNNs and GRI-CNN techniques are intended to improve the CNN capability in terms of system generalization. Depending upon the application, the step angle ($x^o$) may or may not need to be very small in order to create a useful GRI-CNN system for an application. Notice that the requirements of $m \cdot x = 90$ and "m" is an integer must be met to make sure that the GRI-CNN system functions as a "full cycle gearwheel" and each RI-CNN sub-system serves as an "individual gear-tooth". This method would provide a highly systematic GRI-CNN system for applications requiring a greatly stable output result when the input vector is rotated with an arbitrary angle. For simplicity, we only used the second version of RI-CNN-2 for the construction of SNK and GNK in this study. The first version of RI-CNN-2 which combines sub-channels of the Dih4 family at the end of convolution processes would also be able to serve as the core component in both structures. Though its outputs at and after the first flatten layer are the same as those of its second version counterpart, but the back-propagation through these two versions inside a GRI-CNN system would take different routes and is expected to produce somewhat different training results. See Lo et al (2018 (b)) for reference.

Several technical points worth mentioning between an individual RI-CNN and a GRI-CNN system:

1) signals received before the flatten layer are rotationally symmetric for an individual RI-CNN sub-channel with any rotated input vector of the same transformation family. Signals received on and after the first flatten layer are identical.

2) combined signals received at corresponding nodes of the first flatten layer are not the same between RI-CNN sub-channels in GRI-CNN/GSK and GRI-CNN/GNK structures. These signals are networked and used in the classification section (through the MLP or the MCPCNN) to compose final outputs at the end of the GRI-CNN system.

3) combined signals received at nodes of the output layer are the same with any input versions of participated gear-tooth angles in the individual GRI-CNN system.

Based on our initial results, we found that GRI-CNN with GSK and GNK structures performed the best results for all ranges of a pre-set step angle. Since multiple convolution channels are commonly used in a CNN, the use of multiple SSK and GSK channels in a GRI-CNN system would produce similar performance to that of GSK and GNK is anticipated. Based on this initial result, it seems that the step angle set at $1^o$ is appropriate for an input matrix size of 128x128 pixels. Since the final set of trained kernels are less abrupt than their counterparts during the training, we expect the output consistency using final set of trained kernels would be even greater for all 4 GRI-CNN structures. Based on system components and the frame work used in proposed GRI-CNN structures, one can estimate that the computation time required by SSK or SNK is much less than that of GSK or GNK. However, the computation time for training a viable GRI-CNN system may end up the same when more channels of SSK or SNK are used to make up a total number of free parameters needed. This study focused on the consistency of output vector from the proposed GRI-CNN structures. No CNN classification performance study was performed but will be studied and reported in our future studies. Since the GRI-CNN research and applications are in its infancy, our initial results may or may not be applicable to a study subject having certain data characteristics.

Best of all, our study demonstrated that with a proper design a virtually isotropic GRI-CNN can be constructed and systematically evaluated for a given application.